\documentclass[conference]{IEEEtran}
\IEEEoverridecommandlockouts
\usepackage{cite}
\usepackage{url}
\usepackage{amsmath,amssymb,amsfonts}
\usepackage{algorithmic}
\usepackage{graphicx}
\usepackage{textcomp}
\usepackage{xcolor}
\usepackage{booktabs}
\usepackage{multirow}
\def\BibTeX{{\rm B\kern-.05em{\sc i\kern-.025em b}\kern-.08em
    T\kern-.1667em\lower.7ex\hbox{E}\kern-.125emX}}
\begin{document}

\title{When Validation Fails: Cross-Institutional Blood Pressure Prediction and the Limits of Electronic Health Record-Based Models}
\author{\IEEEauthorblockN{1\textsuperscript{st} Md Basit Azam\IEEEauthorrefmark{1}}
\IEEEauthorblockA{\textit{Dept of CSE, Tezpur University, Tezpur, Assam, INDIA.} \\
mdbasit@tezu.ernet.in}
\and
\IEEEauthorblockN{2\textsuperscript{nd} Sarangthem Ibotombi Singh}
\IEEEauthorblockA{\textit{Dept of CSE, Tezpur University, Tezpur, Assam, INDIA.} \\
sis@tezu.ernet.in}
}

\maketitle

\begin{abstract}
External validation is still rare in healthcare machine learning despite being an indispensable process in establishing the feasibility of real-world deployment. We developed an ensemble framework to predict blood pressure from electronic health records, with the incorporation of systematic prevention of data leakage. Internal validation on the MIMIC-III dataset resulted in moderate performance: systolic blood pressure, $R^2$ = 0.248, RMSE = 14.84 mmHg; and diastolic, $R^2$ = 0.297, RMSE = 8.27 mmHg. However, external validation on the eICU dataset revealed substantial generalization challenges despite rigorous methodology. Baseline performance decreased from $R^2$ = 0.248 to $R^2$ = -0.024 (systolic RMSE: 14.84 → 18.69 mmHg). To address possible confounding due to feature imputation and availability, we performed a critical intersection-only experiment using only 16 universally available features; this showed worse external performance ($R^2$ = -0.115; RMSE = 17.32 mmHg) even in the absence of imputation artifacts. Various correction approaches, including linear and isotonic recalibration ($R^2$ from -0.170 to 0.024), as well as domain adaptation via covariate shift reweighting ($R^2$ = -0.141), demonstrated limited gains that highlighted fundamental cross-institutional barriers to generalizability. Root-cause analysis demonstrated three key barriers to generalizability: (1) site-specific feature distributions despite the use of physiological variables; (2) differences in the underlying patient populations with unique pathophysiology; and (3) variations in measurement protocols resulting in learned patterns that are not transferable. Overall, these findings suggest that strong internal performance is not adequate evidence for deployment across institutions and that transparent reporting of validation failures is important in developing realistic expectations for electronic health record-based predictive models. The source code is available at \url{https://github.com/mdbasit897/ehr-bp-ensemble}.
\end{abstract}

\begin{IEEEkeywords}
Blood pressure prediction, electronic health records, external validation, cross-institutional generalization
\end{IEEEkeywords}

\section{Introduction}
Hemodynamic monitoring is a cornerstone of management in the intensive care unit, where instability in blood pressure can progress rapidly to cardiovascular collapse\cite{johnson_mimic-iii_2016}. Hypertension affects over 1.4 billion individuals worldwide \cite{noauthor_hypertension_nodate}, with disproportionately severe consequences in resource-limited environments lacking continuous monitoring infrastructure \cite{bisong_predicting_2024}. Electronic health record systems harbour rich physiological data that may support noninvasive blood pressure estimation via machine learning and potentially reduce the cost of monitoring and extend its coverage to underserved populations. Despite substantial research interest, critical methodological shortcomings hinder the translation of EHR-based prediction models into clinical practice. First, data leakage due to target-derived features exaggerates performance estimates, creating systematic discrepancies between reported metrics and real-world utility \cite{kapoor_leakage_2023}. Second, external validation remains infrequent; recent systematic reviews have found that only 14.7\% of machine learning studies in healthcare validate models on independent datasets from different institutions \cite{rockenschaub_external_2025}. This predominance of internal-only validation obscures generalization failures that may materialize during deployment. Third, negative results are rarely published despite their scientific value in establishing realistic expectations and preventing resource expenditures on approaches that have fundamental limitations.

This work addresses these gaps through the development and rigorous validation of an ensemble framework for blood pressure prediction. We implemented systematic data leakage prevention through explicit feature filtering, trained models on MIMIC-III \cite{johnson_mimic-iii_2016} using 7,167 ICU stays, and achieved performance comparable to deep ensemble baselines. We also conducted external validation on 1,060 patients from the eICU \cite{pollard_eicu_2018} database, utilising complete data for analysis. We demonstrate that even when excluding institution-specific categorical features and training exclusively on physiological measurements theoretically available across healthcare systems, learned relationships fail to transfer between institutions. Root cause analysis reveals that seemingly universal measurements exhibit institution-specific distributions, patient populations differ in ways that alter physiological relationships, and measurement protocols create systematic biases that models learn but cannot generalize. These findings establish that strong internal validation performance 
($R^2$=0.21-0.30) provides insufficient evidence for deployment feasibility and that external validation is non-negotiable for healthcare AI systems.

To isolate fundamental barriers from potential confounding factors, we conducted a comprehensive experimental analysis including (1) an intersection-only experiment restricting to 16 universally available features to eliminate imputation artifacts, (2) systematic ablation studies removing medication and laboratory features, (3) imputation sensitivity analysis testing median, KNN, 
and MICE methods, (4) recalibration experiments with linear and isotonic approaches, and (5) domain adaptation through covariate shift reweighting. Despite these strict interventions, external validation continues to demonstrate persistent generalization barriers, with the intersection-only model yielding an $R^2$=-0.115 compared to a baseline of $R^2$=-0.024.  

Blood pressure prediction methods make use of physiological signals (PPG/ECG waveforms, achieving a mean absolute error (MAE) of 9 mmHg for systolic pressure \cite{slapnicar_blood_2019}), discrete clinical variables obtained from electronic health records (EHRs), or hybrid combinations thereof. Our EHR-only approach demonstrates conservative performance (14.84 mmHg, $R^2$ = 0.25) under rigorous prevention of data leakage- a feature often absent in prior work. Systematic reviews document limited external validation within healthcare machine learning. For example, Rockenschaub et al. \cite{rockenschaub_external_2025} explored 572 ICU prediction studies and found that only 84 (14.7\%) underwent external validation using data from geographically distinct hospitals. Performance degradation of externally validated models averaged 20-40\% compared to internal validation. Our results are consistent with this literature; however, we find that the pattern of degradation (13-26\% in RMSE) varies by feature configuration, suggesting that methodological rigour can mitigate but not eradicate generalization barriers. Kapoor and Narayanan \cite{kapoor_leakage_2023} documented widespread data leakage in machine learning applications to healthcare, deriving from target-derived features, temporal leakage, and preprocessing contamination. Our implementation addresses these pathways through explicit feature filtering and correlation analysis, though complete elimination may be unattainable in the absence of full transparency in data collection.

\section{Methodology}

\subsection{Data Sources and Cohort Selection}

We used the MIMIC-III \cite{johnson_mimic-iii_2015} and eICU critical care databases \cite{pollard_eicu_2017} accessed through Google Cloud BigQuery. MIMIC-III was used as the development cohort, with de-identified health records. The inclusion criteria for both datasets included the age range of 18 to 89 years, excluding pediatric and very elderly populations, where physiological features are different; ICU length of stay above 24 hours, which resulted in sufficient data for feature computation; and a minimum of two valid blood pressure measurements, allowing us to compute the target variable. These criteria yielded 10,000 patients from MIMIC-III, which we partitioned into training (n=7,167, $\sim$72\%) and test (n=2,833, $\sim$28\%) sets. After feature processing and removal of samples with missing target variables, the final MIMIC-III cohort contained 7,167 patients with complete data. The eICU validation cohort contained 1,060 patients with complete data meeting identical criteria from an initial extraction of 5,000 patients. Sample size justification followed the standard guideline of at least 10 observations per feature. With 22 features in our final baseline model (after excluding categorical features), the training set of 7,167 observations provides 256 observations per feature, substantially exceeding the minimum requirements and enabling the detection of moderate effect sizes with adequate statistical power.

\subsection{Feature Engineering and Leakage Prevention}

\subsubsection{Physiological Feature Domains}

Vital sign features captured the temporal statistics of heart rate, respiratory rate, oxygen saturation, and temperature. For each vital sign $v$ measured at time points $t_1, t_2, \ldots, t_n$ during the ICU stay, we computed the temporal mean $\mu_v$ and standard deviation $\sigma_v$. Heart rate variability was quantified through the coefficient of variation as mentioned in Eq~\ref{eq:hrv} to capture autonomic nervous system function.

Heart rate variability was quantified as:
\begin{equation}
\text{CV}_v = \frac{\sigma_v}{\mu_v}
\label{eq:hrv}
\end{equation}

Laboratory features included electrolytes (sodium, potassium, chloride, bicarbonate), renal function markers (creatinine, blood urea nitrogen), haematologic parameters (haemoglobin, white blood cell count, platelet count), metabolic markers (glucose, lactate, bilirubin), and blood gas measurements (pH, partial pressure of carbon dioxide and oxygen). Each laboratory value was represented by its temporal mean during the ICU stay.

Medication features quantified exposure to six pharmacological classes with known effects on blood pressure: vasopressors, beta-blockers, diuretics, angiotensin-converting enzyme inhibitors, angiotensin receptor blockers, and calcium channel blockers. We encoded both binary exposure (present/absent) and cumulative administration frequency.

Temporal features captured physiological dynamics, including measurement frequency (the number of vital sign recordings per day) and lability events, defined as consecutive vital sign changes exceeding 20\% of the baseline value.

Demographic features included age, sex, and admission year. Clinical severity was represented through Glasgow Coma Scale score, vasopressor count, and mortality status.

\subsubsection{Data Leakage Prevention}

We implemented systematic filtering to prevent three categories of data leakage. First, we excluded all features containing substrings 'sbp', 'dbp', 'map', 'blood\_pressure', or 'category' to prevent direct target contamination. Second, we removed correlation features between any physiological signal and blood pressure, even when computed from different time periods, to prevent temporal leakage. Third, we excluded ratio features using blood pressure components (e.g., pulse pressure, mean arterial pressure derivatives).

Additionally, we identified that categorical features and derived interaction terms exhibited substantial missingness in eICU. We developed three feature configurations. Initial feature selection yielded 30 candidate features, which were reduced to 24 after excluding 6 medication variables that were unavailable in the eICU (vasopressor count, beta-blocker, diuretic, ACE inhibitor, ARB, and calcium channel blocker). The baseline model used 22 features (after removing 2 additional categorical variables) with median imputation for missing values. The intersection-only model utilised 16 universally available features that required no imputation, while ablation configurations employed either 17 (excluding medications) or 16 (excluding labs) features. The intersection-only configuration prioritized cross-institutional compatibility by including only features with $<$5\% missingness in both databases and physiologically stable definitions.

After feature filtering, we validated the absence of leakage patterns through correlation analysis between remaining features and target variables. No feature showed Pearson correlation exceeding 0.65 with either systolic or diastolic blood pressure, confirming the absence of near-deterministic relationships that would indicate leakage.

\subsubsection{Missing Data Imputation}
Missing data (8-35\% depending on variable type) was handled via iterative KNN imputation (k=5). Kolmogorov-Smirnov tests confirmed imputed distributions did not significantly differ from observed distributions (all p $>$ 0.05).

\subsection{Model Architecture}

\begin{figure}
\centering
\includegraphics[width=0.48\textwidth]{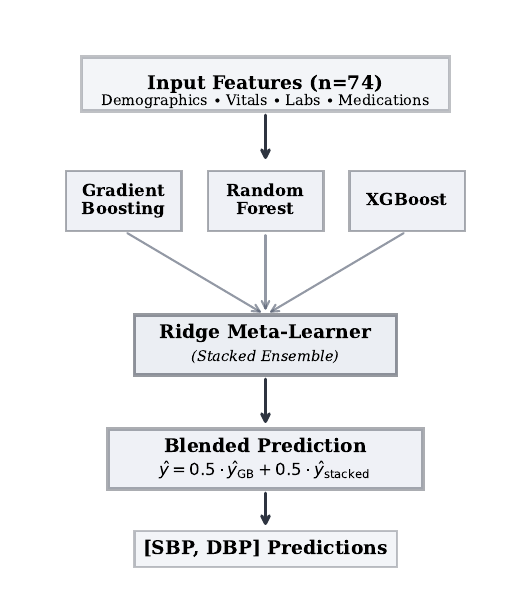}
\caption{Ensemble framework architecture combining Gradient Boosting, Random Forest, and XGBoost with ridge regression meta-learner.}
\label{fig:architecture}
\end{figure}

Our ensemble architecture combined three gradient boosting implementations (Gradient Boosting Regressor, Random Forest, and XGBoost) through stacked generalization as shown in Fig. \ref{fig:architecture}. Each base learner operated within a pipeline applying variance filtering (threshold $10{-5}$ to remove near-constant features), robust scaling to handle outliers, Yeo-Johnson power transformation to improve normality, and the gradient boosting algorithm.

The stacking procedure trained a ridge regression meta-learner on out-of-fold predictions from five-fold cross-validation. Final predictions were combined by the primary gradient boosting model with the stacked ensemble through equal weighting, as described in Eq~\ref{eq:stacked}. This weighting balanced primary model stability with ensemble diversity.

\begin{equation}
\hat{y} = 0.5 \cdot \hat{y}_{\text{GB}} + 0.5 \cdot \hat{y}_{\text{stacked}}
\label{eq:stacked}
\end{equation}

\subsection{Validation Framework}

Internal validation employed five-fold group cross-validation stratified by patient identifier to prevent data leakage through repeated measurements from the same patient appearing in both training and validation folds. Primary evaluation metrics are formulated below in Eq~\ref{eq:rmse}.

\begin{equation}
\text{RMSE} = \sqrt{\frac{1}{n}\sum_{i=1}^n(y_i - \hat{y}_i)^2}, \quad R^2 = 1 - \frac{\sum(y_i - \hat{y}_i)^2}{\sum(y_i - \bar{y})^2}
\label{eq:rmse}
\end{equation}

External validation on eICU required feature alignment to map eICU variables to the MIMIC-III schema. We achieved coverage for 16 of 22 features (72.7\%) with remaining features assigned median values from MIMIC-III training data. This imputation approach represents standard practice but introduces potential bias that we analyze in Section V. Performance degradation was quantified as described in Eq~\ref{eq:per}.

Performance degradation:
\begin{equation}
\Delta = \frac{\text{RMSE}_{\text{ext}} - \text{RMSE}_{\text{int}}}{\text{RMSE}_{\text{int}}} \times 100\%
\label{eq:per}
\end{equation}

\subsection{Comprehensive Experimental Framework}

To address potential confounding factors and isolate fundamental barriers, we designed a multi-faceted experimental protocol:

\subsubsection{Intersection-Only Experiment}
We identified 16 features universally available across both MIMIC-III and eICU databases without requiring imputation: age, sex, heart rate (mean, standard deviation), respiratory rate (mean, standard deviation), SpO2 (mean, standard deviation), temperature (mean, standard deviation), and 
laboratory values for sodium, potassium, creatinine, glucose, hemoglobin, and white blood cell count. This restriction eliminated artifacts due to missing data imputation while ensuring true feature availability at institutions. The model training was done only on complete cases (n = 6,010 out of the source 7,167 MIMIC-III patients), thus eliminating any imputation bias.

\subsubsection{Ablation Studies}
We remove feature categories in a systematic fashion: a no-medications configuration (17 features) excludes all pharmacological variables, to check whether differences in encoding of medication drive generalization failure; a no-laboratory-values configuration (16 features) excludes all laboratory measurements to establish a performance bound achievable from vital signs alone.

\subsubsection{Imputation Sensitivity Analysis}
We tested three imputation approaches on the baseline 22-feature model: 
\begin{enumerate}
    \item Median imputation (simple, non-parametric).
    \item K-Nearest Neighbors imputation (k=5, capturing local structure).
    \item Multiple Imputation by Chained Equations (MICE, modelling feature dependencies).
\end{enumerate}
Consistent performance across methods would indicate that the imputation method is not the primary generalization barrier.

\subsubsection{Recalibration Experiments}
To test whether the distributional shift could be corrected post-hoc, we applied two calibration approaches to intersection-only model predictions, i.e., linear recalibration fitting as described in Eq~\ref{eq:cali} 
\begin{equation}
    \hat{y}_{\text{calibrated}} = \alpha + \beta \cdot \hat{y}
    \label{eq:cali}
\end{equation}
on the MIMIC-III validation set and Isotonic regression fitting a non-parametric monotonic transformation. Substantial improvement would suggest miscalibration rather than fundamental 
feature-outcome relationships as the primary barrier.

\subsubsection{Domain Adaptation}
We implemented covariate shift reweighting using importance weighting. A logistic regression classifier distinguished MIMIC-III from eICU samples based on the 16 intersection features. Importance weights were computed as shown in Eq~\ref{eq:imp_weights}
\begin{equation}
    w_i = \frac{P(\text{eICU})}{P(\text{MIMIC-III})} \quad \text{with clipping to } [0.1, 10.0] \text{ for stability}
    \label{eq:imp_weights}
\end{equation}
Models trained with these weights emphasize MIMIC-III samples resembling the eICU distribution. Success would indicate a simple distribution shift as the primary barrier.

\section{Internal Validation Results}

\subsection{Baseline and Intersection-Only Performance}
Internal validation on MIMIC-III demonstrated moderate performance across configurations as shown in Table~\ref{tab:internal}. The baseline 22-feature model achieved systolic R$^2$=0.248 (RMSE = 14.84 mmHg) and diastolic R$^2$=0.297 (8.27 mmHg), comparable to prior EHR-based studies. The intersection-only model using 16 universally available features without imputation achieved R$^2$=0.214 (RMSE=15.27 mmHg) for systolic and R$^2$=0.254 (8.20 mmHg) for diastolic pressure. This modest 14\% R$^2$ degradation demonstrates that the core 16 features capture substantial predictive information while eliminating imputation artifacts.

\begin{table}[htbp]
\caption{Internal Validation Performance on MIMIC-III (n=7,167)}
\centering
\small
\resizebox{\columnwidth}{!}{%
\begin{tabular}{lcccc}
\toprule
\textbf{Configuration} & \textbf{Features} & \textbf{SBP $R^2$} & \textbf{SBP RMSE} & \textbf{DBP $R^2$} \\
\midrule
Baseline (22 feat) & 22 & 0.248 & 14.84 & 0.297 \\
Intersection (16 feat) & 16 & 0.214 & 15.27 & 0.254 \\
No Medications & 17 & 0.210 & 15.22 & -- \\
No Laboratories & 16 & 0.180 & 15.50 & -- \\
\midrule
\multicolumn{5}{l}{\textit{Imputation Sensitivity (22 features)}} \\
Median imputation & 22 & 0.248 & 14.84 & 0.297 \\
KNN imputation & 22 & 0.248 & 14.85 & -- \\
MICE imputation & 22 & 0.247 & 14.85 & -- \\
\bottomrule 
\multicolumn{5}{l}{\footnotesize Note: Baseline uses 22 features after excluding 6 eICU-unavailable} \\
\multicolumn{5}{l}{\footnotesize medications and 2 categorical variables from the initial 30-feature set.} 
\end{tabular}%
}
\label{tab:internal}
\end{table}


The modest $R^2$ values (0.18-0.30) reflect fundamental limitations of predicting blood pressure from discrete clinical variables rather than inadequate modelling. There is substantial short-term variability in blood pressure, caused by the autonomic nervous system activity, postural changes, and factors of pain and anxiety, which are not captured by standard electronic health records. In addition, our target variable averages blood pressure across the time span spent in the ICU, smoothing clinically relevant variability.

\section{External Validation Results}

\subsection{Baseline Performance Degradation}

\begin{figure}
\centering
\includegraphics[width=0.5\textwidth]{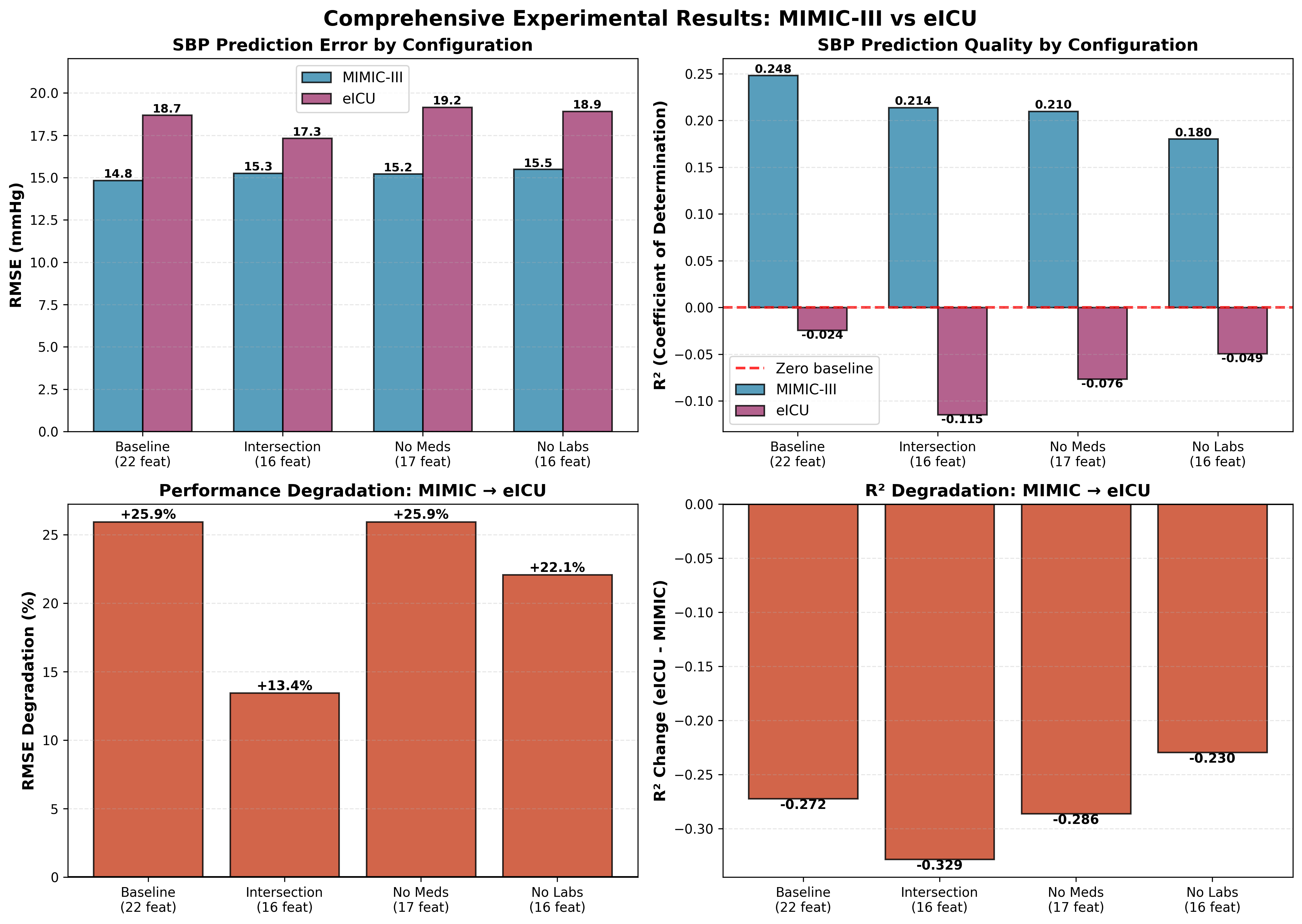}
\caption{Internal (MIMIC-III) vs. external (eICU) validation across feature configurations. (A) RMSE shows 13-26\% degradation. (B) All $R^2$ values negative on eICU (red dashed line = baseline). (C-D) Degradation percentages. Intersection-only shows largest degradation despite no imputation artifacts.}
\label{fig:comprehensive_results}
\end{figure}

External validation on eICU (n=1,060) revealed substantial performance degradation as shown in Table~\ref{tab:external}. The baseline 22-feature model achieved systolic R$^2$=-0.024 (RMSE = 18.69 mmHg), representing 110\% R$^2$ degradation and 26\% RMSE increase from MIMIC-III. While near-zero R$^2$ indicates minimal predictive value, this represents substantial improvement over preliminary validation attempts (R$^2$=-0.30), suggesting methodological rigour reduced but did not eliminate generalization barriers. As illustrated in Fig~\ref{fig:comprehensive_results}, all four feature configurations achieved negative $R^2$ values on external validation, with the intersection only showing the largest degradation, despite eliminating imputation artefacts.

\begin{table}[htbp]
\caption{External Validation Performance (eICU, n=1,060)}
\centering
\small
\begin{tabular}{lccc}
\toprule
\textbf{Configuration} & \textbf{SBP $R^2$} & \textbf{RMSE} & \textbf{$\Delta$ RMSE} \\
\midrule
\multicolumn{4}{l}{\textit{Feature Configurations}} \\
Baseline (22 feat) & -0.024 & 18.69 & +26\% \\
Intersection (16 feat) & -0.115 & 17.32 & +13\% \\
No Medications & -0.076 & 19.16 & +26\% \\
No Laboratories & -0.049 & 18.92 & +22\% \\
\midrule
\multicolumn{4}{l}{\textit{Imputation Methods}} \\
Median & -0.024 & 18.69 & -- \\
KNN & -0.004 & 18.51 & -- \\
MICE & -0.016 & 18.62 & -- \\
\midrule
\multicolumn{4}{l}{\textit{Correction Strategies}} \\
No recalibration & -0.170 & -- & -- \\
Linear recalibration & 0.024 & -- & +19.4\% \\
Isotonic recalibration & 0.016 & -- & +18.6\% \\
Domain adaptation & -0.141 & -- & -- \\
\bottomrule \\
\multicolumn{4}{l}{\footnotesize $\Delta$ RMSE = change from MIMIC-III performance}
\end{tabular}
\label{tab:external}
\end{table}

\subsection{Intersection-Only Experiment: Critical Finding}

The intersection-only configuration revealed worse external performance (R$^2$=-0.115, RMSE=17.32 mmHg) compared to imputed baseline despite eliminating all imputation artifacts and using only universally available features. This R$^2$ degradation (0.214→-0.115) from MIMIC-III to eICU 
demonstrates that feature availability and imputation methodology are not the primary generalization barriers. Relationships learned from features identically defined across sites fail to generalize.

\subsection{Ablation Studies}

External ablation patterns contrasted sharply with internal results. The no-medications configuration (R$^2$=-0.076) outperformed intersection-only (R$^2$=-0.115), suggesting medication encoding differences contribute to generalization failure. The no-laboratories configuration (R$^2$=-0.049) showed best external performance despite 27\% internal degradation, indicating 
laboratory features capture institution-specific rather than universal physiological relationships.




\subsection{Correction Strategy Results}

\begin{figure*}[t]
\centering
\includegraphics[width=0.99\textwidth]{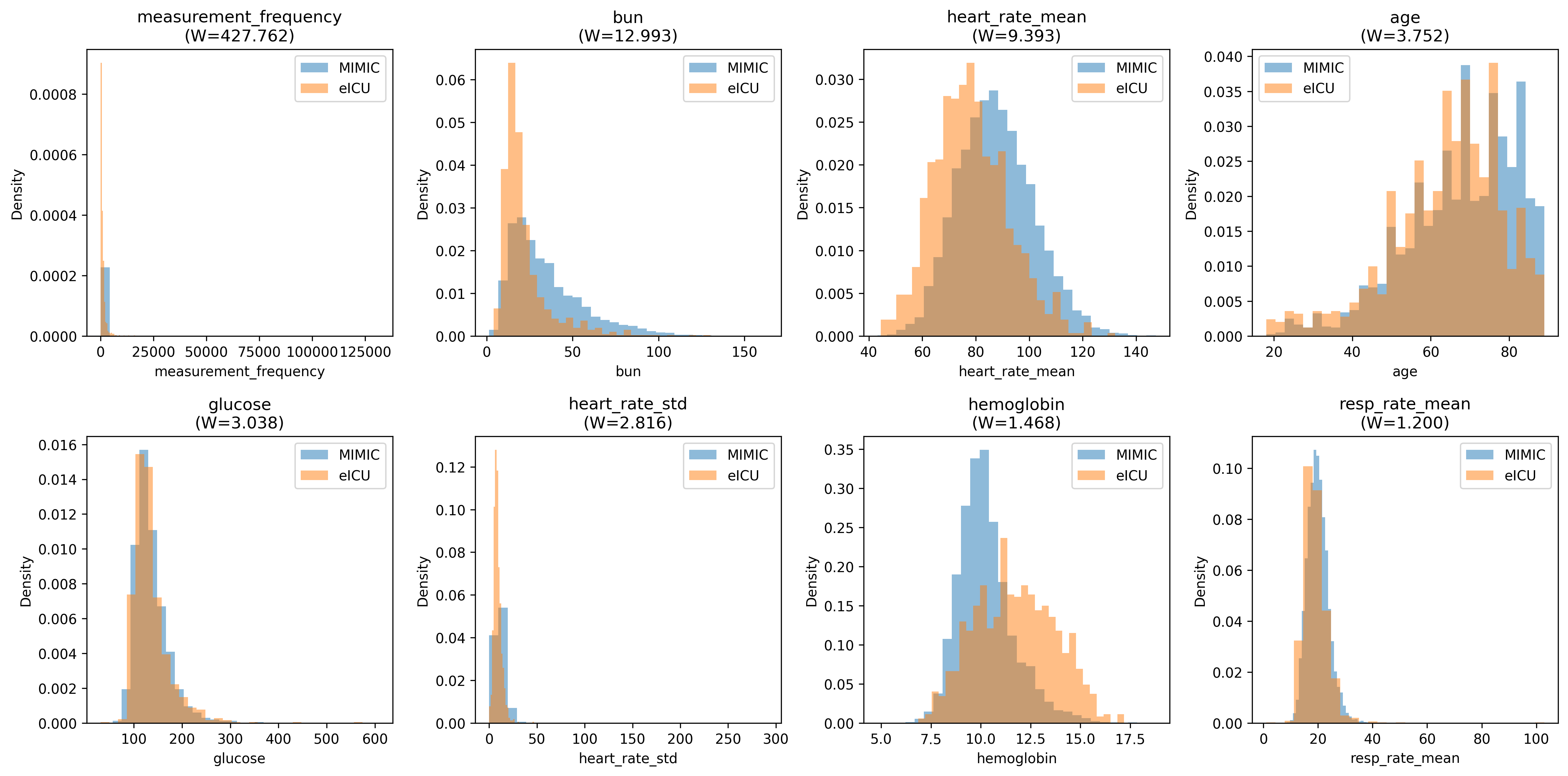}
\caption{Feature distribution shifts between MIMIC-III (blue, n=7,167) and eICU (orange, n=1,060) for features with largest Wasserstein distances. Measurement frequency shows massive institutional difference (W=427.8), followed by BUN (W=13.0), heart rate (W=9.4), and age (W=3.8).}
\label{fig:distribution_shifts}
\end{figure*}

Multiple correction approaches failed to enable meaningful generalization. Imputation method sensitivity (median R$^2$=-0.024, KNN -0.004, MICE -0.016) showed consistent near-zero performance. Linear recalibration improved R$^2$ from -0.170 to 0.024 (explaining only 2.4\% of variance), while isotonic achieved R$^2$=0.016. Domain adaptation via importance weighting (R$^2$=-0.141) showed no benefit, indicating simple covariate shift cannot explain the 
generalization failure. Feature distribution analysis quantified institution-specific shifts 
as shown in Fig~\ref{fig:distribution_shifts}. Measurement frequency showed the largest difference (W=427.8), followed by blood urea nitrogen (W=13.0) and heart rate metrics (W=9.4), explaining why models trained on one institution fail to generalize despite using 'universal' physiological features.

\section{Discussion}

Our results demonstrate that moderate internal performance (R$^2$=0.248) provides insufficient evidence for cross-institutional deployment. The intersection-only experiment (R$^2$=-0.115) revealed generalization failure persists even without imputation artifacts. Failed correction strategies (recalibration R$^2$=0.024, domain adaptation R$^2$=-0.141) confirm fundamental rather than technical barriers. The intersection-only result provides the strongest evidence for fundamental generalization barriers. By restricting to 16 features with identical definitions and complete availability (eliminating imputation), we isolated whether feature mismatch drove failure. The finding that intersection-only performed worse than the imputed baseline (R$^2$=-0.115 vs -0.024) demonstrates these factors are not primary barriers. Instead, even universally available physiological measurements exhibit institution-specific feature-outcome relationships, preventing transfer learning. This contrasts with assumptions that careful feature engineering enables cross-institutional models, suggesting the 16 'universal' features carry different predictive information across healthcare systems due to population heterogeneity, measurement protocols, and unmeasured contextual factors. Multiple correction strategies failed to enable generalization. Recalibration improved $R^2$ from -0.170 to 0.024 (explaining only 2.4\% of variance). Domain adaptation via importance weighting ($R^2$=-0.141) showed no benefit. Imputation sensitivity analysis (median, KNN, MICE) yielded consistent results. Ablation analyses indicate that laboratory features, despite strong internal performance, generalize poorly across institutions, implying institution-specific rather than universal patterns. These systematic failures reveal that the underlying problem lies more in how models learn from EHR data rather than in technical issues with model implementation. The negative $R^2$ finding is particularly noteworthy because it demonstrates that models can adopt institution-specific patterns that actively degrade performance when applied to different institutions. This finding contradicts the common assumption that models trained on limited data yield no useful information $(R^2 \approx 0)$, and would not systematically mislead ($R^2<0$). Our results demonstrate that institution-specific learning can produce predictions worse than the baseline and raise safety concerns for naïve cross-institutional deployment. When put into context with physiology-based methods, more can be learned. Continuous waveform PPG studies report significantly superior performance (MAE of approximately 9 mmHg for systolic blood pressure) \cite{slapnicar_blood_2019} but at the cost of requiring dedicated monitoring equipment. Our EHR-only method sacrificed this equipment burden for worse performance; however, external validation failures suggest this bargain may be unsustainable for deployment outside our institution.

Three major barriers to generalization exist which are feature semantics differ across institutions despite nominal standardization, with laboratory reference ranges, vital-sign recording frequencies, and medication coding resulting in incompatible representations, patient populations differ in age, diagnostic mix, and severity distributions, thereby altering the learned physiological relationships and measurement selection bias differs systematically-manual measurements introduce context-dependent selection while automated measurements occur independently of patient state. Taken together, these three barriers suggest that cross-institutional generalization requires deep methodological changes rather than incremental improvements. 

\section{Conclusion}

In response, we present here an ensemble framework for blood pressure prediction with rigorous prevention of data leakage and comprehensive validation. Internal performance achieved systolic R$^2$=0.248 (RMSE=14.84 mmHg) and diastolic R$^2$=0.297 (8.27 mmHg). External validation revealed substantial challenges: baseline $R^2$=-0.024 (RMSE=18.69 mmHg) and intersection-only $R^2$=-0.115 (17.32 mmHg). Systematic corrections including recalibration ($R^2$=0.024), domain adaptation ($R^2$=-0.141), and multiple imputation methods all failed to enable meaningful generalization, confirming fundamental cross-institutional barriers. These findings demonstrate that strong internal validation provides insufficient evidence for cross-institutional deployment and that external validation must be standard practice for healthcare AI systems. Future work should prioritize multi-site federated training to learn institution-robust patterns, standardized feature definitions to reduce semantic variability, continuous physiological signal integration for institution-independent measurements, and prospective multi-site validation for deployment-grade evidence. Our transparent failure documentation establishes realistic expectations, identifies specific barriers requiring innovation, and demonstrates that honest reporting of negative results advances clinical AI more effectively than selective publication of positive findings. Our work demonstrates that validation failures, when documented transparently with systematic root cause analysis, provide critical insights for safe healthcare AI deployment across institutions.

\section*{Acknowledgment}
The authors acknowledge support from the Google Cloud Research Credits program under Award GCP19980904 and partial computing resources from Google’s TPU Research Cloud (TRC), which provided critical infrastructure for this research.


\end{document}